\documentclass[preprint,10pt]{elsarticle}

\AtBeginDocument{%
  \def\thefnote{\myfnsymbol{fnote}}}
\makeatletter
\def\myfnsymbol#1{\expandafter\@myfnsymbol\csname c@#1\endcsname}
\def\@myfnsymbol#1{\ifcase #1\or $\dagger$\or $\#$\else \@ctrerr\fi}
\def\fntext[#1]#2{\g@addto@macro\@fnotes{%
   \refstepcounter{fnote}\elsLabel{#1}%
   \def\thefootnote{\thefnote}
   \global\setcounter{footnote}{\c@fnote}%
   \footnotetext{#2}}}
\makeatother

\usepackage{parskip}

\usepackage{amssymb}

\usepackage{makecell}

\usepackage[table]{xcolor}

\usepackage{tikz}

\usepackage{pifont}

\usepackage{enumitem}

\usepackage{placeins}

\usepackage{arydshln}

\usepackage{threeparttable}

\usepackage{hyperref}
\hypersetup{
    colorlinks=true,
    linkcolor=blue,
    filecolor=magenta,      
    urlcolor=cyan,
    pdftitle={Overleaf Example},
    pdfpagemode=FullScreen,
    }

\renewcommand{\appendixname}{Supplementary}

\usepackage[left=3.5cm, right=3.5cm]{geometry}

\usepackage{caption}
\DeclareCaptionFont{tenpt}{\fontsize{9}{11}\selectfont}
\usepackage{etoolbox}
\AtBeginEnvironment{tabular}{\fontsize{9}{11}\selectfont}
\AtBeginEnvironment{tabularx}{\fontsize{9}{11}\selectfont}
\AtBeginEnvironment{longtable}{\fontsize{9}{11}\selectfont}

\usepackage{fancyhdr}
\journal{arXiv}

\begin{document}

\begin{frontmatter}



\title{\textbf{Circle of Willis Centerline Graphs: A Dataset and Baseline Algorithm}}


\author[zhaw,dqbm]{Fabio Musio}
\author[zhaw]{Norman Juchler}
\author[dqbm]{Kaiyuan Yang}
\author[dqbm]{Suprosanna Shit}
\author[dqbm]{Chinmay Prabhakar}
\author[dqbm]{Bjoern Menze}
\author[zhaw]{Sven Hirsch}

\affiliation[zhaw]{organization={Institute of Computational Life Sciences, Zurich University of Applied Sciences}, 
            city={Waedenswil},
            country={Switzerland}}

\affiliation[dqbm]{organization={Department of Quantitative Biomedicine, University of Zurich},
            city={Zurich},
            country={Switzerland}}

\address{\{fabio.musio, sven.hirsch\}@zhaw.ch}

\begin{abstract}
The Circle of Willis (CoW) is a critical network of arteries in the brain, often implicated in cerebrovascular pathologies. Voxel-level segmentation is an important first step toward an automated CoW assessment, but a full quantitative analysis requires centerline representations. However, conventional skeletonization techniques often struggle to extract reliable centerlines due to the CoW’s complex geometry, and publicly available centerline datasets remain scarce.
To address these challenges, we used a thinning-based skeletonization algorithm to extract and curate centerline graphs and morphometric features from the TopCoW dataset, which includes 200 stroke patients, each imaged with magnetic resonance angiography (MRA) and computed tomography angiography (CTA). The curated graphs were used to develop a baseline algorithm for centerline and feature extraction, combining U-Net-based skeletonization with A* graph connection. Performance was evaluated on a held-out test set, focusing on anatomical accuracy and feature robustness.
Further, we used the extracted features to predict the frequency of fetal PCA variants, confirm theoretical bifurcation optimality relations, and detect subtle modality differences between MRA and CTA. The baseline algorithm consistently reconstructed graph topology with high accuracy (F1 = 1), and the average Euclidean node distance between reference and predicted graphs was below one voxel. Features such as segment radius, length, and bifurcation ratios showed strong robustness, with median relative errors below 5\% and Pearson correlations above 0.95.
Our results demonstrate the utility of learning-based skeletonization combined with graph connection for anatomically plausible centerline extraction. We emphasize the importance of going beyond simple voxel-based measures by evaluating anatomical accuracy and feature robustness. The dataset and baseline algorithm have been released to support further method development and clinical research.\\
\end{abstract}





\end{frontmatter}

\section{Introduction}
\label{sec:introduction}
The Circle of Willis (CoW) is an important system of arteries that connects the anterior and posterior circulations of the brain, as well as the left and right cerebral hemispheres \citep{osborn2013osborn}. Due to its central role in cerebral blood flow, the CoW is believed to play an important role in cerebrovascular pathologies such as aneurysms and stroke \citep{hoksbergen2003absent, van2015completeness, kim2016BMJ, rinaldo2016relationship, hindenes2023anatomical}.
Despite its clinical importance, assessing CoW anatomy remains a manual expert task. The CoW comprises multiple small, branching vessels, and anatomical variants, such as hypoplastic or absent segments, are common \citep{krabbe1998circle, iqbal2013comprehensive}. This complexity and variability underscore the need for automated and reliable tools for CoW analysis.

Therefore, we organized the challenge `topology-aware anatomical segmentation of the Circle of Willis for CTA and MRA', or TopCoW for short, in 2023 and 2024 \citep{yang2023benchmarking}, framing CoW characterization as a multiclass anatomical segmentation task. The challenge dataset \citep{topcow_challenge_organizers_2025_15692630} comprises 200 stroke patients imaged with both magnetic resonance angiography (MRA) and computed tomography angiography (CTA), with voxel-level annotations for 13 cerebral arteries.
TopCoW led to crowd-sourced segmentation algorithms that represent the current state-of-the-art and demonstrated the utility of these algorithms in downstream clinical tasks such as CoW variant classification and aneurysm localization.

While voxel-level segmentation is a crucial first step, it is insufficient for a full assessment of the CoW. Morphometric features such as vessel radius, tortuosity, and bifurcation angles are essential for quantitative analysis of the cerebral vasculature, as they support hemodynamic modeling, enable detection of anatomical variants such as fetal PCA, and facilitate studies of population-level variability of the CoW. Extracting these features typically requires a centerline representation. However, extracting accurate centerlines from segmentation masks remains challenging. Classical skeletonization techniques \citep{saha2016survey} -- such as morphological thinning, distance transforms, and Voronoi-based methods -- are used in established vascular analysis frameworks \citep{chen2018development, izzo2018vascular}. Nonetheless, these methods often struggle with complex vascular geometries and rely on extensive post-processing to correct spurious branches and maintain anatomical plausibility.\\
Recent advances in convolutional neural networks (CNN) have led to a new class of learned skeletonization methods that treat centerline extraction as a segmentation problem \citep{panichev2019u, nguyen2021u, vargas2025skelite}. These approaches have shown promising results, as CNNs learn hierarchical spatial features directly from image data, handling complex vascular geometries more robustly than rule-based methods and reducing the need for post-processing. 

Despite the importance of centerline representations for vascular modeling, publicly available vessel centerline datasets remain scarce. The CASILab healthy MRA database \citep{bullitt2005vessel} provides brain artery centerlines for a subset of healthy subjects but lacks detailed anatomical labeling of the CoW. The CROWN Challenge dataset \citep{vos2025evaluation, R05G1L_2023} includes verified morphometric features of the CoW such as vessel diameters and bifurcation angles, but no centerline representations. Other datasets are either limited to specific regions like the carotid bifurcation \citep{eulzer2023fully, eulzer_2024_10695923}, derived from mouse brain data \citep{paetzold2021whole}, focused on other anatomical regions like the retina \citep{imed_2020_12778666}, or are synthetic in nature \citep{tetteh2020deepvesselnet, batten_2025_15742395}. \\
To fill this gap, we introduce the first publicly available CoW centerline graph dataset that includes anatomically labeled vessel segments, spatial descriptions of key anatomical nodes, and a rich set of morphometric features. We believe that such a dataset can facilitate research in many fields: Hemodynamic modeling and blood flow simulation, automated detection of bifurcation points, learning-based skeletonization, as well as quantitative studies of CoW anatomical variability.\\
Building on recent advances in deep learning-based skeletonization, we also propose an end-to-end pipeline that generates anatomically labeled centerline graphs directly from TopCoW segmentation masks. Our baseline method combines the nnU-Net framework \citep{isensee2023nnUNet}, trained on our curated centerline dataset, with the A* pathfinding algorithm \citep{hart1968formal} to generate anatomically correct centerlines and enable reliable morphometric feature extraction. To assess its performance, we benchmark the pipeline against test data from our centerline dataset, which was generated using a traditional topological thinning-based algorithm \citep{drees2021scalable}. The evaluation focuses not only on segmentation quality but also on anatomical accuracy and feature robustness, both of which are critical for clinical applications.
 
We summarize our main contributions as follows:
\begin{enumerate}
    \item \textbf{A novel CoW centerline graph dataset:} We present the first publicly available dataset with anatomically labeled centerlines and morphometric features for the CoW, supporting a wide range of research applications.
    \item \textbf{A baseline algorithm for centerline and feature extraction:} We introduce a U-Net-based approach for skeletonization, combined with the A* algorithm to obtain anatomically accurate centerlines and features. The code complements existing TopCoW models \citep{topcow_challenge_organizers_2024_15665435}, enabling an end-to-end pipeline from raw angiographic images to quantitative vascular analysis.
    \item \textbf{A systematic comparison of centerline-derived features:} We evaluate the consistency of morphometric features extracted from two different centerline sources, providing insights into feature robustness. This analysis informs the selection of reliable features for downstream clinical and research applications.
\end{enumerate}

\section{Material and Methods}
\label{sec:method}
\subsection{TopCoW Dataset}
The TopCoW cohort includes 200 stroke patients admitted to the University Hospital Zurich (USZ) between 2018 and 2019. The median age was 74 years (interquartile range: 60.8–82.3), with 59.5\% male patients. 
The dataset is randomly split into 125 training, 5 validation, and 70 test cases. Each patient has both a CTA and an MRA, resulting in a total of 400 images. Across both modalites, all images were annotated using 13 artery segment labels, as shown in Figure \ref{fig:cow_segments_bifurcations}(a)–(b). For details on data acquisition, inclusion criteria, preprocessing, and data annotation, see \citep{yang2023benchmarking}.\\
The verified TopCoW multiclass masks of the CoW form the basis for our centerline extraction, algorithm development and evaluation. Raw imaging data and multiclass masks for the training set are publicly available via the TopCoW Zenodo dataset repository \citep{topcow_challenge_organizers_2025_15692630}.

\subsection{Voreen for Centerline Graph Extraction}
\label{subsec:voreen_graph_extraction}
\begin{figure}[!ht]
    \centering
    \includegraphics[width=\linewidth]{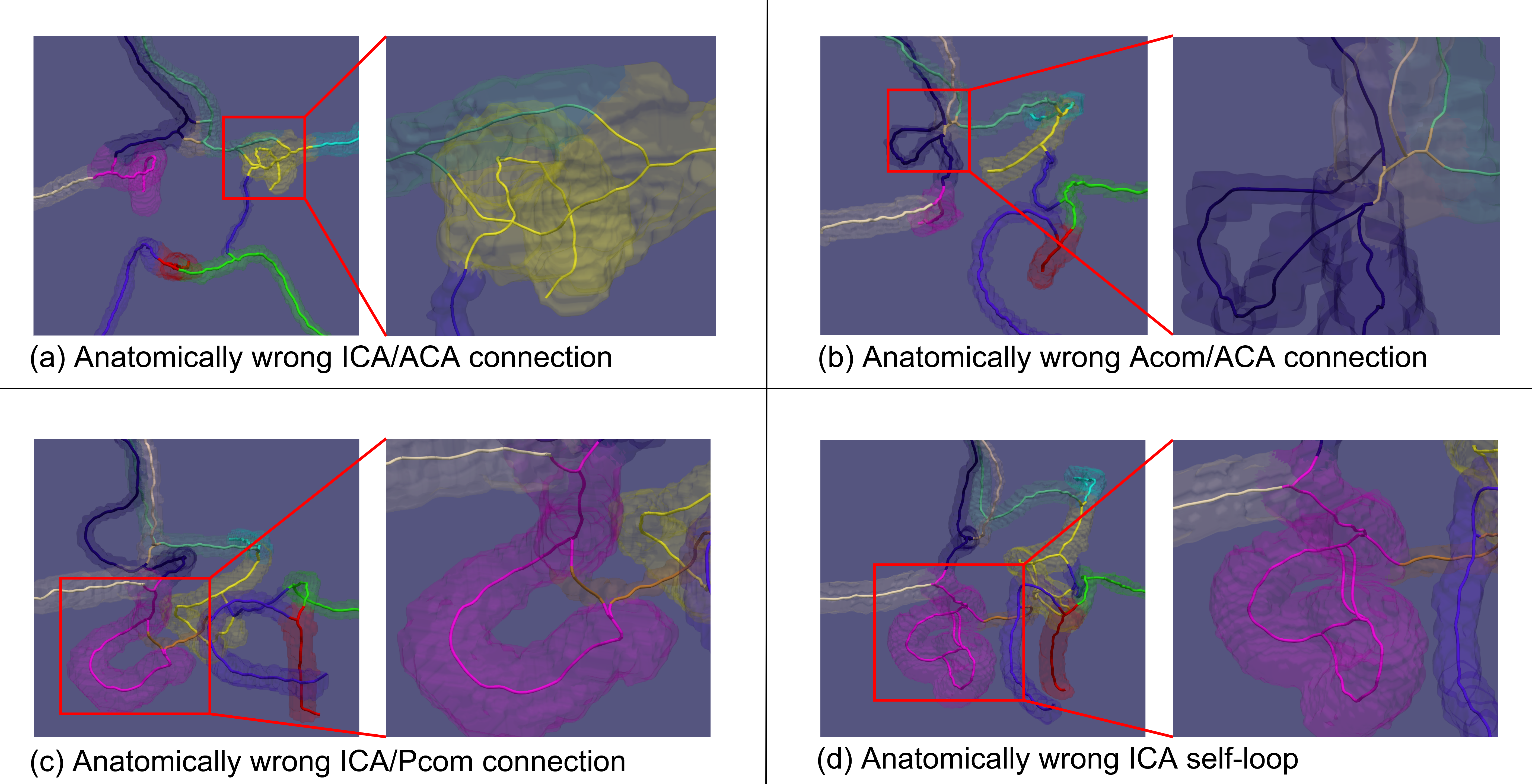}
    \caption{Topological errors of the centerline graphs as introduced by the Voreen skeletonization algorithm: a) Invalid connection between R-ICA and R-ACA, b) invalid connection between Acom and L-ACA, c) invalid connection between L-ICA and L-Pcom, d) self-loop of L-ICA.}
    \label{fig:topological_errors_graphs}
\end{figure}

The centerline extraction is done on the TopCoW masks using the method presented in \citep{drees2021scalable}. One of its key advantages is that it is scalable, robust and deterministic. It has, for example, been successfully applied to whole mouse brain vasculature \citep{paetzold2021whole}.
An implementation of the pipeline is publicly available in \textit{Voreen} (Volume Rendering Engine), which is a framework for the visualization and analysis of multi-modal volumetric data sets \citep{meyer2009voreen}.\\
While effective, this method often introduces topological errors in complex vascular structures, such as invalid connections, loops or extra branches. Figure \ref{fig:topological_errors_graphs} shows some common artifacts as introduced by the Voreen skeletonization tool. To correct these, we apply a series of rule-based and anatomy-informed post-processing steps, including node extraction, selective centerline segment merging, and spurious edge removal. To guarantee anatomical faithfulness, all the processed centerlines were inspected by an expert trained on the CoW anatomy. A detailed description of the post-processing pipeline and Voreen parameter settings is provided in \ref{appendix:graph_extraction_details}.

\subsection{CoW Feature Extraction}
\label{subsec:feature_extraction}
The output of the Voreen graph extraction is a centerline graph consisting of a set of nodes and edges which are enriched with attributes, including spatial features provided by the Voreen tool such as node coordinates and average edge radii. To obtain radius measurements that don't rely on the Voreen tool and can be used independently in our baseline algorithm, we perform a cross-section analysis to estimate edge-wise radii along the centerline. Specifically, we compute the circle-equivalent (CE) radius based on the cross-sectional area. Given its close agreement with Voreen estimates (see \ref{appendix:radius_comparison}), we use the CE radius in all subsequent morphometric analyses for consistency with our baseline method.

\begin{figure}[!ht]
    \centering
    \includegraphics[width=1\textwidth]{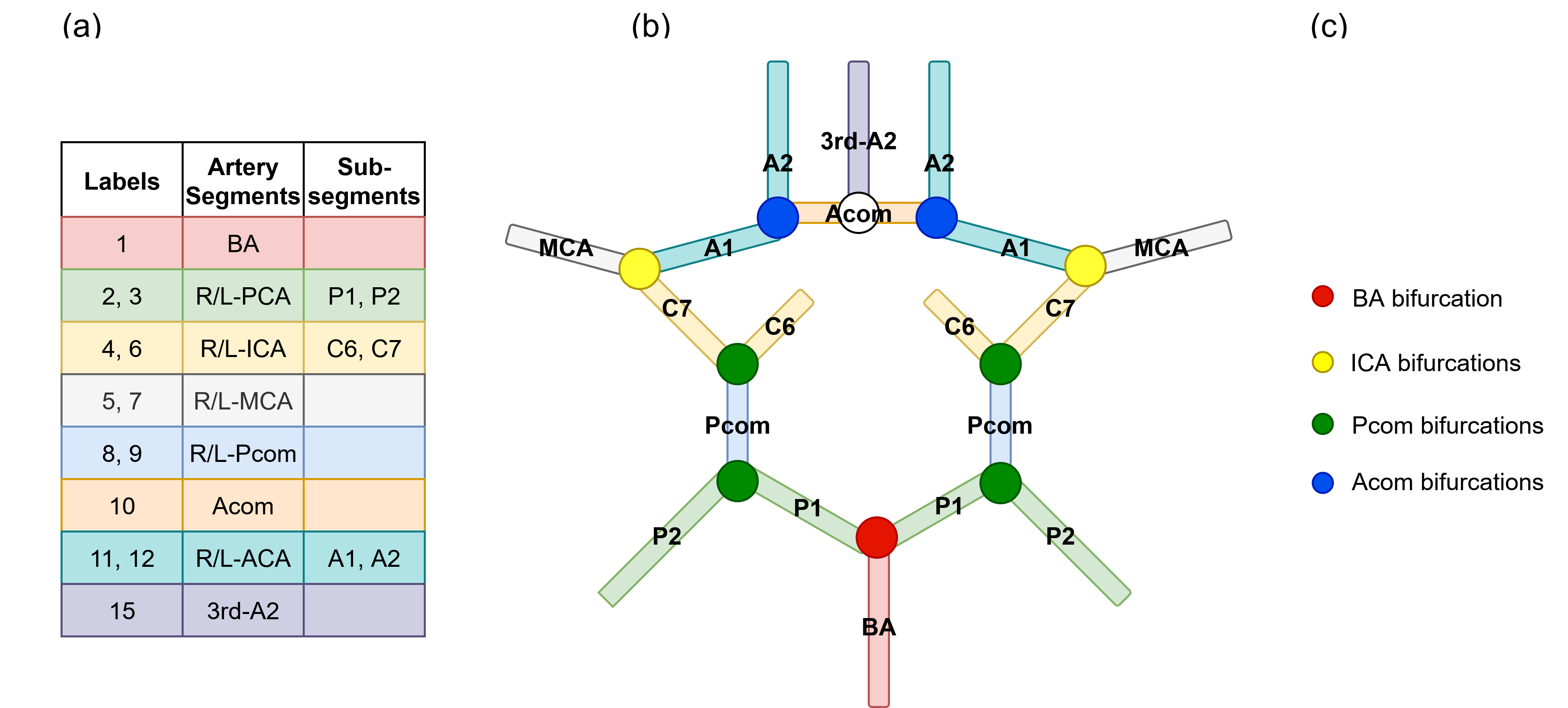}
    \caption{Overview of CoW components used for feature computation: (a) List of artery segments and subsegments with labels as defined by the TopCoW challenge: basilar artery (BA), left and right posterior cerebral artery (PCA), left and right internal carotid artery (ICA), left and right middle cerebral artery (MCA), left and right posterior communicating artery (Pcom), anterior communicating artery (Acom), and left and right anterior cerebral artery (ACA). (b) Schematic representation of the CoW and its components. (c) Anatomical locations of CoW bifurcations.}
    \label{fig:cow_segments_bifurcations}
\end{figure}

While node and edge attributes describe local graph properties, our goal is to derive global geometric descriptors of the CoW. To this end, we subdivide the centerline graph into subsets based on the anatomical segments and bifurcations of the CoW, as shown in Figure \ref{fig:cow_segments_bifurcations}. Based on this subdivision, we compute a comprehensive set of morphometric features for each segment and bifurcation separately, as described below. Additional technical details on the feature extraction process are provided in Appendix \ref{appendix:morphometric_feature_extraction}.

\subsubsection{Segment Features}
\label{subsubsec:segment_features}
For each vessel segment and subsegment depicted in Figure \ref{fig:cow_segments_bifurcations}(a)-(b), we extract the following geometric features: radius, length, tortuosity, volume, and curvature. 
Tortuosity is hereby defined as 
\begin{equation}
    \tau = \frac{C}{L}-1
\end{equation}
where $C$ is the length of the segment and $L$ is the Euclidean distance between its ends.
Segment length, tortuosity, and curvature are computed by fitting smooth cubic splines to the corresponding subgraphs using the \textit{Splipy} library \citep{johannessen2020splipy}. 

\subsubsection{Bifurcation Features}
\label{subsubsec:bif_features}
Figure \ref{fig:cow_segments_bifurcations}(b)–(c) shows the bifurcations considered for analysis. We distinguish between major bifurcations - the BA and two ICA bifurcations - and minor bifurcations, which include two Acom and four Pcom bifurcations.\\
For all the bifurcations, we extract the three bifurcation angles \citep{wischgoll2010modeling}.
For the major bifurcations only, we additionally compute radius- and area-based metrics. These include individual ratios between the radii of the parent and child vessels, the radius and area sum ratios, as well as the bifurcation exponent. The radius sum ratio is defined as the parent vessel radius divided by the sum of the child vessel radii. According to the empirical Finet formula for vascular bifurcations \citep{finet2008fractal}, this ratio is expected to follow:
\begin{equation}
\label{eq:finet_relation}
    r_p = 0.678\cdot(r_{c1} + r_{c2})
\end{equation}
where $r_p$, $r_{c1}$, $r_{c2}$ denote the radii of the parent and the two child vessels, respectively. Similarly, the area sum ratio is defined as the square of the parent radius divided by the sum of the squared child radii.\\
Finally, the bifurcation exponent is the exponent $x$ that satisfies
\begin{equation}
\label{eq:bif_exponent}
    r_p^x=r_{c1}^x+r_{c2}^x.
\end{equation}
This exponent characterizes the relationship between the diameters of the three vessels at a bifurcation. As summarized in Table 1 of \citep{huo2012optimal}, several theoretical models describe optimal branching behavior. One such model, proposed by Huo and Kassab \citep{huo2009scaling}, is based on the minimum energy hypothesis and suggests: 
\begin{equation}
\label{eq:hk_relation}
    r_p^{\frac{7}{3}}=r_{c1}^{\frac{7}{3}}+r_{c2}^{\frac{7}{3}}
\end{equation}
which implies that the power required to transport blood through the bifurcation is minimized.\\
For minor bifurcations (Acom and Pcom), we exclude radius-based features due to the lack of a clearly defined parent-child flow hierarchy and an increased unreliability of radius estimates in the communicating arteries, particularly the Acom. 

\subsection{Baseline Algorithm for Centerline Extraction} 
\label{subsec:baseline_algorithm}
\begin{figure}[!ht]
    \centering
    \includegraphics[width=1\textwidth]{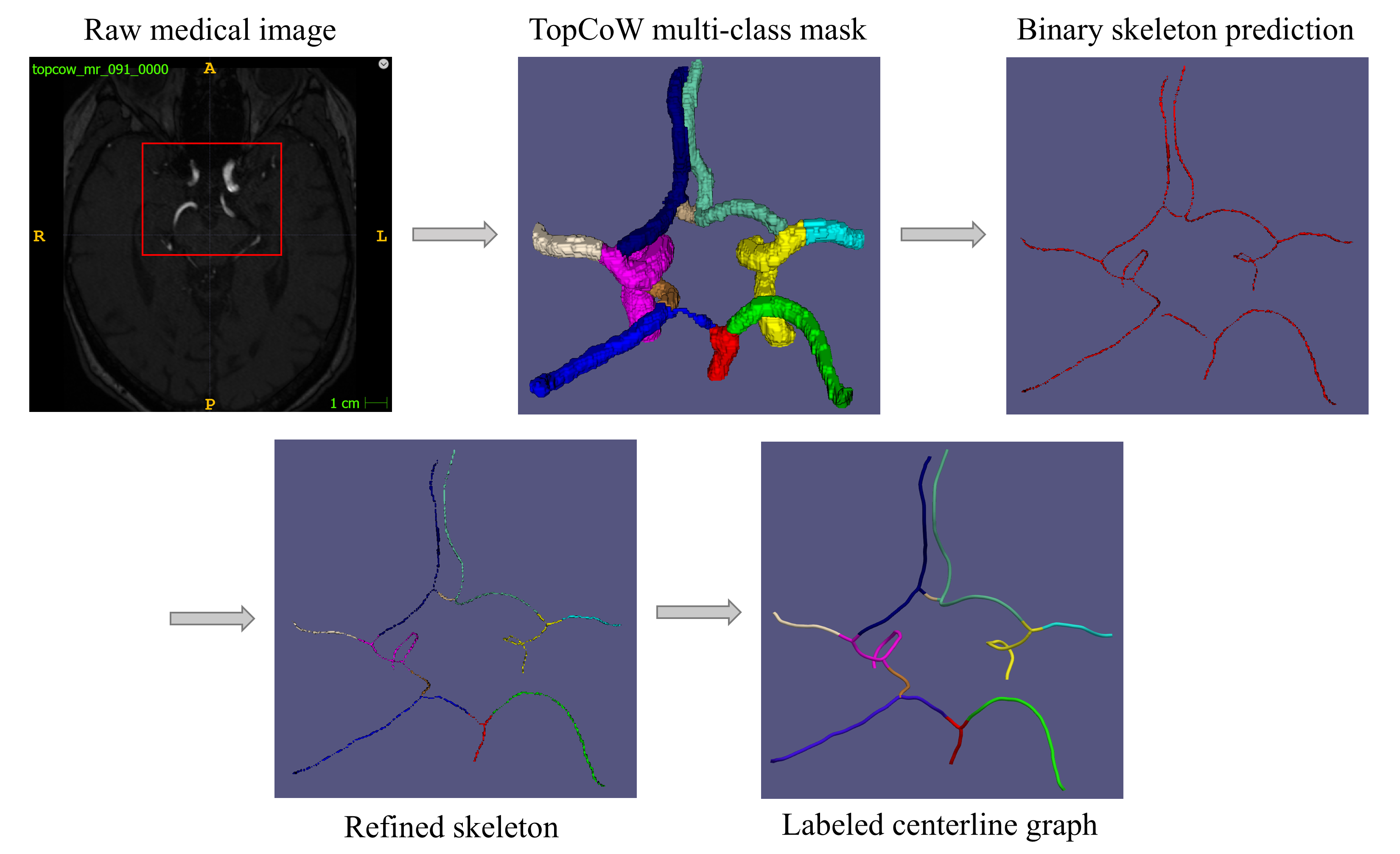}
    \caption{Overview of the baseline centerline extraction pipeline. Starting from a multi-class segmentation mask of the CoW, a U-Net model predicts a binary skeleton. The skeleton is then refined by assigning anatomical labels from the original mask and by reconnecting broken segments using the A$^*$ algorithm. Finally, a labeled centerline graph is extracted using the Voreen tool.}
    \label{fig:cnt_pipeline}
\end{figure}

To address the limitations of conventional skeletonization methods -- particularly their tendency to introduce topological errors -- we reformulate the CoW centerline extraction as a segmentation task and use a learning-based pipeline. The pipeline consists of three stages: (1) skeletonization using a U-Net model with the TopCoW multi-class mask as input and a binary skeleton as output, (2) reconnecting the skeleton using the A* algorithm, and (3) converting the connected skeleton into a graph using the Voreen tool. An overview of the pipeline is shown in Figure \ref{fig:cnt_pipeline}.

A key challenge in this segmentation-based approach is the extreme class imbalance between foreground (skeleton) and background voxels, which leads to over-thinning and fragmented predictions. To mitigate this, we train the U-Net using a combination of Focal Loss \citep{lin2017focal}, which imposes a higher penalty on misclassified foreground voxels, and Tversky Loss \citep{tversky1977features, salehi2017tversky}, with a higher penalty on false negatives. The A$^*$ pathfinding algorithm for reconnecting the predicted skeleton is guided by a custom heuristic cost function that balances minimal Euclidean distance and maximal distance from the segment boundary. This allows us to restore anatomical continuity while avoiding spurious shortcuts. Finally, we convert the voxel-based skeleton into a centerline graph using the Voreen tool, similar to the procedure described in Section \ref{subsec:voreen_graph_extraction}, but using the skeleton as input instead of the full vessel mask. A post-processing step is then applied to the extracted graphs to remove artifacts, extract anatomically relevant nodes, and smooth the vessel segments. Further details on the implementation of the pipeline are provided in \ref{appendix:baseline_implementation_details}.

\section{Results}
\label{section:results}
\subsection{CoW Centerline Dataset}
\label{subsec:dataset}
\begin{figure}[!ht]
    \centering
    \includegraphics[width=1\textwidth]{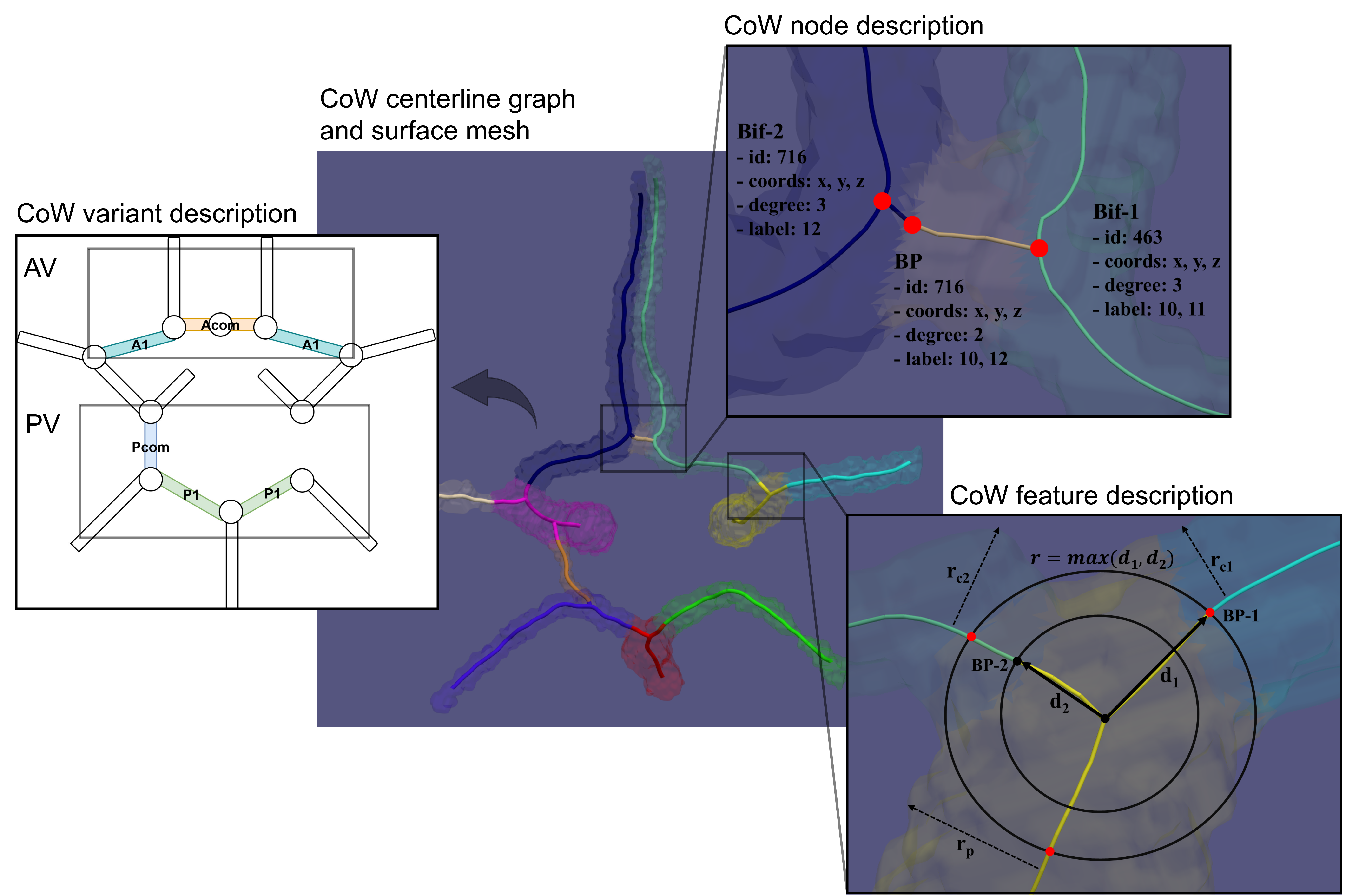}
    \caption{CoW Centerline Dataset: The dataset includes centerline graphs and surface meshes enriched with spatial, topological, and semantic node and cell attributes. Three derived data files provide complementary information: the variant file encodes anterior (AV) and posterior (PV) CoW variants based on the presence or absence of four artery segments each, along with fetal-type PCA variants; the node file lists key anatomical nodes with ID, degree, label, and coordinates; and the feature file contains morphometric features for CoW segments and bifurcations.}
    \label{fig:cnt_dataset}
\end{figure}

The dataset comprises 400 CoW centerline graphs from 200 patients, derived from the TopCoW 2024 dataset. 
On average, graphs extracted from CTA-derived masks contain $843\pm125$ nodes and $842\pm126$ edges, while those extracted from MRA-derived masks contain $931\pm105$ nodes and $931\pm106$ edges.\\
For each patient and imaging modality, the dataset includes an attributed centerline graph and surface mesh --- both stored as VTK PolyData objects --- as well as three files containing derived data, as summarized in Figure \ref{fig:cnt_dataset}. Table \ref{table:graph_features} lists all attributes of the centerline graphs, including topological, spatial and semantic features. \\
The derived data consists of variant, node and feature information:
The variant file describes the presence or absence of four anterior segments: L-A1, Acom, 3rd-A2, and R-A1; four posterior segments: L-Pcom, L-P1, R-P1, and R-Pcom; as well as fetal-type variants \citep{shaban2013circle}: fetal L-PCA and fetal R-PCA. It also includes annotations for arterial fenestrations \citep{cooke2014cerebral}, which may occur in five segments: L-A1, Acom, R-A1, L-P1, and R-P1. More details on the CoW variant classification are provided in \ref{appendix:cow_variant_classification}.
The node file contains anatomically relevant graph nodes, including start and end points (typically of degree 1), bifurcation points (degree 3), and segment boundary points (degree 2). Each node entry includes an ID, degree, segment label and spatial location. A comprehensive list of possible nodes per segment is provided in \ref{appendix:graph_post_processing}, Table \ref{table:nodes_per_segment}.
Finally, the feature file contains morphometric descriptors for each segment and bifurcation, as described in Section \ref{subsec:feature_extraction}. 

\begin{table*}[!ht]
\caption{Overview of the attributes included in the CoW centerline graphs. Node degree is a topological property reflecting graph connectivity. 3D coordinates and voreen\textunderscore radius are spatial features extracted using the Voreen tool, while mis\textunderscore radius (maximally inscribed sphere) and ce\textunderscore radius (circle-equivalent) are spatial features computed separately on cross-sections along the centerline. The label attribute is semantic, identifying the underlying CoW anatomy. For each attribute, the table lists its data type and the corresponding vtkPolyData method used for access.}
\label{table:graph_features}
\centering
    \begin{threeparttable}
    \begin{tabular}{l|l|l|l|l}
        \hline
        \multicolumn{5}{c}{\bfseries Graph attribute overview}\\
        \hline
        \bfseries Name & \makecell[l]{\bfseries Attribute\\\bfseries type} & \makecell[l]{\bfseries vtkPolyData\\\bfseries method} & \makecell[l]{\bfseries Data\\\bfseries type} & \bfseries Description \\
        \hline
        degree & node & .GetPointData() & int & Node degree \\
        coords & node & .GetPoints() & tuple & 3-Tuple (x, y, z) \\
        \cdashline{1-5}
        labels & edge & .GetCellData() & int & TopCoW segment labels\\
        ce\textunderscore radius\tnote{*} & edge & .GetCellData() &  double & Circle-equivalent radius \\
        mis\textunderscore radius & edge & .GetCellData() &  double & Max. inscribed sphere radius \\
        voreen\textunderscore radius & edge & .GetCellData() &  double & Voreen average radius \\
        \hline
    \end{tabular}
    \begin{tablenotes}
    \footnotesize
    \item[*] Note that ce\textunderscore radius is convertible to the vessel cross-sectional area through $A_{cs}=r_{ce}^2\pi$.
    \end{tablenotes}
    \end{threeparttable}
\end{table*}

\subsection{CoW Vascular Morphology of the TopCoW Cohort} 
Three key findings are summarized below. More detailed results on variant distributions, segment and bifurcation features are provided in \ref{appendix:topcow_characteristics}.\\
1) \textit{Anatomical variants are common:} Complete CoW configurations --- defined by the presence of the Acom, both A1s, P1s, and Pcoms --- are relatively uncommon, observed in only 13\% of CTA and 18\% of MRA cases. The Pcoms are the most frequently missing segments, with bilateral absence observed in 41\% (CTA) and 40\% (MRA) of cases, and unilateral absence in an additional 33.5\% (CTA) and 31.5\% (MRA). The 3rd-A2 (ACA trifurcation) segment is identified in 10.5\% of CTA and 11.5\% of MRA cases, which is on the higher end but still within the reported prevalence range of 2–13\% \citep{dimmick2009normal}.
Fetal-type PCA variants are automatically identified comparing the radii of the Pcom and P1 segments (see \ref{appendix:cow_variant_classification}). Across the full TopCoW dataset, fetal variants are present in 27\% (CTA) and 24.5\% (MRA) of patients --- unilaterally in 19\% (CTA) and 19.5\% (MRA), and bilaterally in 8\% (CTA) and 5\% (MRA).\\
2) \textit{Modality-dependent morphometric differences:} MRA consistently yields larger vessel radii than CTA across all segments except the Acom (see supplementary Table \ref{table:cow_segment_features}). This difference is statistically significant (Wilcoxon signed-rank test, $p<0.05$), with the largest discrepancy observed in the P1 segment (mean paired difference: 0.26 mm). Vessel segments are also consistently found to be absent more frequently in CTA than in MRA (see supplementary Table \ref{table:missing_segments_overall}). These differences suggest a modality-dependent bias in the TopCoW annotations, likely reflecting inherent differences in vessel depiction between the modalities. \\
3) \textit{Bifurcation symmetry and optimality patterns:} Quantitative analysis of bifurcation radius ratios and exponents is summarized in Table \ref{table:cow_bifurcation_ratio_features}. At the BA bifurcation, we observe the expected symmetry between the left and right PCAs, reflected in a child-to-child radius ratio close to 1.
In contrast, the ICA bifurcations consistently show asymmetry with the MCA having a 15–20\% larger radius. 
Regarding the optimal bifurcation diameter relationship, the MRA data shows strong agreement with both the Finet formula (Equation \ref{eq:finet_relation}) and the theoretical bifurcation exponent of $\frac{7}{3}$ (Equation \ref{eq:hk_relation}). These findings align well with the results reported in \citep{huo2012optimal}. In contrast, the CTA data deviates more substantially from this optimal branching pattern, with the bifurcation exponent in particular being significantly smaller than the theoretical value. This suggests a relative underestimation of child vessel calibers compared to the parent vessel in CTA --- most notably at the BA bifurcation, and to a lesser extent at the ICA bifurcations. 
 
 \begin{table*}[!ht]
\caption{Median (Q1-Q3) values of bifurcation radius features across all major CoW bifurcations, reported separately for CTA and MRA. The table lists the radius sum ratio (motivated by Finet's formula \ref{eq:finet_relation}), the radius ratio of the child vessels, and the bifurcation exponent $x$ that satisfies Equation \ref{eq:bif_exponent}. For each bifurcation listed in row order, the parent vessel (p) and child vessels (c1, c2) are as follows: (p) BA, ICA; (c1) R-PCA, MCA; (c2) L-PCA, ACA.}
\label{table:cow_bifurcation_ratio_features}
\centering
    \begin{tabular}{l|l|l|l|l}
        \hline
        \multicolumn{5}{c}{\bfseries CTA bifurcation radius features} \\
        \hline
        \bfseries Bifurcation & \bfseries Support & \makecell[l]{\bfseries Ratio(p,c1+c2)} & \makecell[l]{\bfseries Ratio(c1,c2)} & \makecell[l]{\bfseries Bifurcation\\\bfseries exponent} \\
        \hline
        BA & 173 & $0.72$ $(0.68-0.80)$ & $0.98$ $(0.88-1.10)$ & $1.90$ $(1.45-2.30)$\\
        ICA & 386 & $0.69$ $(0.64-0.75)$ & $1.20$ $(1.06-1.33)$ & $2.15$ $(1.70-2.70)$\\
        \hline
        \multicolumn{5}{c}{}\\
        \multicolumn{5}{c}{\bfseries MRA bifurcation radius features}\\
        \hline
        \bfseries Bifurcations & \bfseries Support & \makecell[l]{\bfseries Ratio(p,c1+c2)} & \makecell[l]{\bfseries Ratio(c1,c2)} & \makecell[l]{\bfseries Bifurcation\\\bfseries exponent} \\
        \hline
        BA & 181 & $0.66$ $(0.61-0.72)$ & $0.99$ $(0.90-1.08)$ & $2.30$ $(1.85-3.10)$\\
        ICA & 387 & $0.67$ $(0.63-0.72)$ & $1.16$ $(1.05-1.28)$ & $2.35$ $(1.85-3.05)$\\
        \hline
    \end{tabular}
\end{table*}

\subsection{Evaluation of the Baseline Algortihm for Centerline Extraction}
The baseline algorithm was evaluated on the TopCoW test set using the manually verified Voreen centerline graphs as reference (see Section \ref{subsec:voreen_graph_extraction}). The full pipeline --- from segmentation to feature extraction --- runs in $81\pm14$s per case on a ThinkPad P1 with RTX A2000 GPU. nnU-Net inference with a 5-fold ensemble takes $23\pm3$s, and skeleton connection adds $16\pm6$s.

\subsubsection{U-Net Skeletonization Performance}
Skeletonization performance is summarized in Table \ref{table:unet_seg_performance}, both before and after connecting the skeleton. We adopt evaluation metrics similar to those in \citep{vargas2025skelite}: the Dice similarity coefficient serves as the overlap measure; topological correctness is assessed via connected component errors (errors in the 0th Betti number); and thickness is evaluated on the graphs extracted from the skeletons by computing the average and 99th percentile radius.\\
Dice scores and thickness estimates remain stable before and after skeleton connection, while topological errors are effectively eliminated by reducing the $\beta_0$ error to near zero. Average thickness values are close to the resolution limit of 1 voxel (0.25mm). Compared to prior work \citep{vargas2025skelite}, our method achieves competitive results, with Dice scores and topological correctness at the upper end of previously reported values, despite longer inference times due to heavy upsampling and ensembling.

\begin{table*}[!ht]
\caption{Mean $\pm$ standard deviation of the Dice score, 0th Betti number error, and average and maximum (99th percentile) thickness for the U-Net predicted binary skeletons and the connected skeletons on the TopCoW test set (n=140). No distinction is made between skeletons predicted from CTA-derived masks and those from MRA-derived masks.}
\label{table:unet_seg_performance}
\centering
    \begin{tabular}{l|l|l|l|l}
        \hline
        \multicolumn{5}{c}{\makecell[c]{\bfseries Skeletonization performance (n=140)}} \\
        \hline
        \bfseries Skeleton data & \makecell[l]{\bfseries Dice\\\bfseries score [\%]} & \bfseries $\beta_0$ error & \makecell[l]{\bfseries Average\\\bfseries thickness [mm]} & \makecell[l]{\bfseries 99th max\\\bfseries thickness [mm]} \\
        \hline
        U-Net prediction & $55.38\pm2.43$ & $5.31\pm3.53$ & $0.24\pm0.01$ & $0.35\pm0.02$ \\
        Connected & $55.35\pm2.45$ & $0.01\pm0.08$ & $0.23\pm0.01$ & $0.35\pm0.02$ \\
        \hline
    \end{tabular}
\end{table*}

\subsubsection{Comparison of Derived CoW Quantities}
\label{subsubsec:cow_comparison}
Based on the skeleton predicted by the U-Net, the pipeline generates centerline graphs, surface meshes, and corresponding CoW variant, node, and feature data. The primary objective is to obtain a reliable quantitative representation of the CoW, as captured in these derived quantities. Ideally, different methods would yield identical results that perfectly reflect the true underlying anatomy. In practice, however, discrepancies arise, which are quantified in the following.\\
Variant classification matches the reference data: Across all binary variables --- indicating the presence or absence of anterior and posterior segments, fetal variants and fenestrations --- the micro-averaged F1 score was 1.0. This suggests that the graph topology is accurately reconstructed by the skeletonization and connection algorithm, which was the primary motivation and objective of the baseline method.\\
Node localization accuracy was assessed via Euclidean distance to reference nodes. The results are shown in Table \ref{table:avg_node_dist} for different node types. Average deviations are approximately 1 voxel (0.25mm), with bifurcation nodes showing slightly higher errors due to their anatomical complexity. These values reflect the spatial uncertainty inherent in the centerline-based node extraction.

\begin{table*}[!ht]
\caption{Mean $\pm$ standard deviation of the Euclidean distances between predicted and reference CoW nodes on the TopCoW test set (n=140). Distances are reported separately for major bifurcation nodes, minor bifurcation nodes, and boundary nodes. No distinction is made between nodes extracted from CTA-derived masks and those from MRA-derived masks.}
\label{table:avg_node_dist}
\centering
    \begin{tabular}{l|l|l}
        \hline
        \multicolumn{3}{c}{\makecell[c]{\bfseries Average node distances}} \\
        \hline 
        \bfseries Nodes & \bfseries Support & \bfseries avg. distance [mm] \\
        \hline
        Major bifurcations & 399 & $0.280 \pm 0.156$ \\
        Minor bifurcations & 423 & $0.286 \pm 0.219$ \\
        Boundary points & 2529 & $0.207 \pm 0.188$ \\
        \cdashline{1-3}
        Overall & 3351 & $0.226 \pm 0.192$ \\
        \hline
    \end{tabular}
\end{table*}

\begin{figure}[!ht]
    \centering
    \includegraphics[width=1\textwidth]{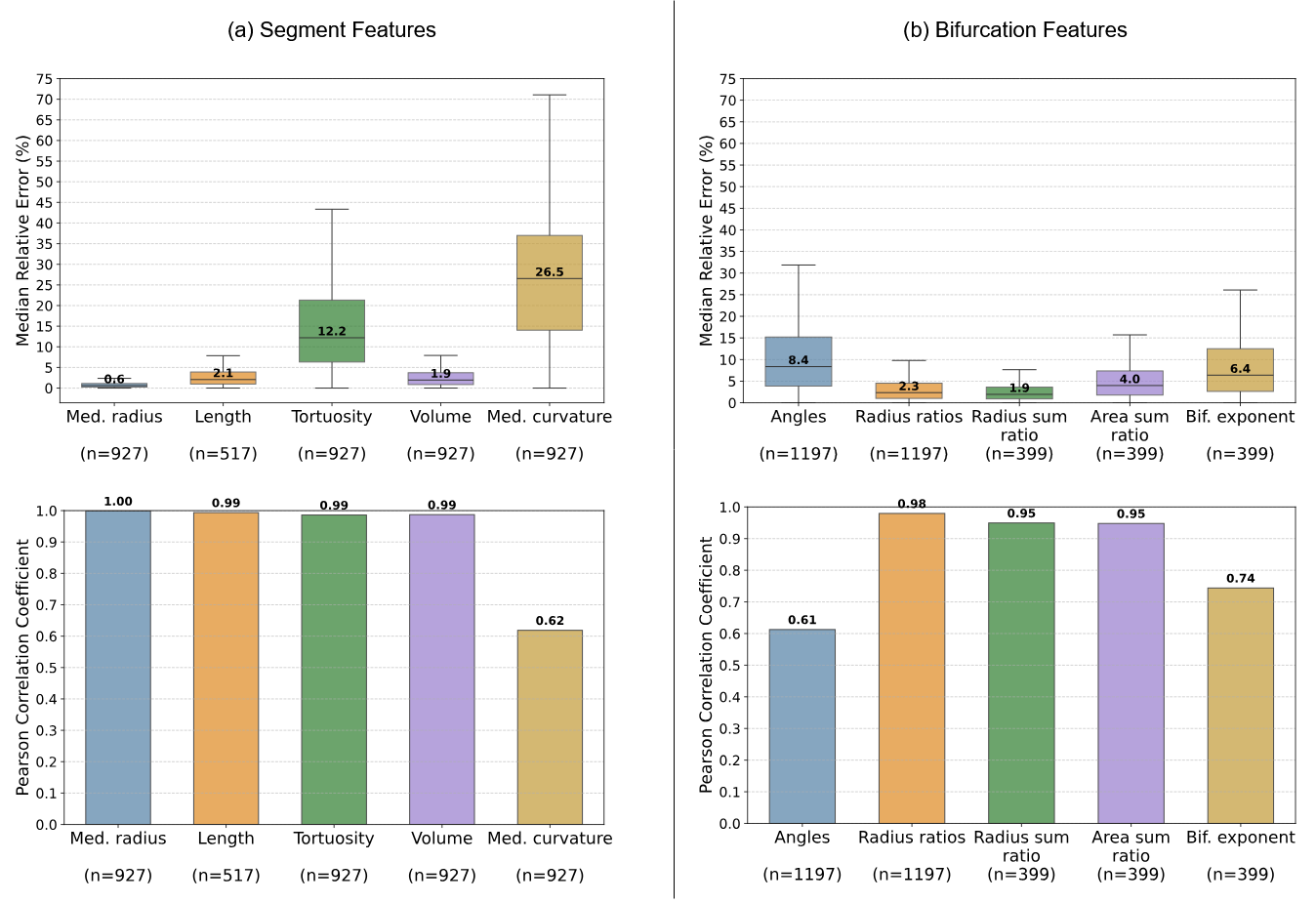}
    \caption{Comparison of morphometric features extracted from the Voreen-based and U-Net–predicted centerline graphs, evaluated by median relative error (MedRE) and Pearson correlation coefficient. (a) comparison of segment features, aggregated across the P1, C7, Pcom, and A1 segments. The Acom segment is excluded due to its significantly higher error rates. (b) comparison of the bifurcation features, aggregated across the BA and both ICA bifurcations.}
    \label{fig:feature_comparison}
\end{figure}

To assess the reliability of segment and bifurcation features extracted from the U-Net–predicted centerlines, we compared them to those derived from the Voreen-based reference graphs. Figure \ref{fig:feature_comparison} shows the median relative error (MedRE) and Pearson correlation coefficients (Pearson r) for (a) the segment features, and (b) the bifurcation features. While clinically validated ground truth values for these morphometric CoW features are unavailable and costly to obtain, this comparison offers insight into their robustness across different centerline extraction methods.\\
For the segment features, we restrict our attention to segments that are located strictly within the CoW. The features shown in Figure \ref{fig:feature_comparison}(a) are aggregated across the P1, C7, Pcom, and A1 segments. 
Radius, length, and volume estimates show very good agreement (median MedRE $\leq$ 2.1\%, Pearson r $\geq$ 0.99). Tortuosity retains strong correlation despite higher MedRE, suggesting that while absolute values may deviate, the relative trends are preserved, still making it a useful feature. Curvature exhibits the highest MedRE and comparatively low correlation, indicating limited reliability.
The Acom segment was excluded from the analysis due to notably higher MedRE values across most features, reflecting lower reliability caused by its small, globular shape and less well-defined centerline. 
A more detailed comparison for each individual segment is provided in 
\ref{appendix:baseline_feature_comparison}, Table \ref{table:segment_feature_comparison}. \\
The bifurcation features shown in Figure \ref{fig:feature_comparison}(b) are aggregated across the major CoW bifurcations, specifically the BA and ICA bifurcations.
Radius and area ratios demonstrate strong agreement between methods (median MedRE $\leq$ 4.0\%, Pearson r $\geq$ 0.95), indicating high consistency across methods.
Angle and exponent features show moderate median MedRE ($\leq$ 8.4\%) and moderate correlation ($\leq$ 0.74), suggesting that these features capture general bifurcation trends but may be less reliable for precise quantitative analysis. Sampling points farther from the bifurcation point may improve robustness, particularly for angle measurements. Overall, these features may still be suitable for downstream use, depending on the application, though their limitations should be carefully considered.
A detailed comparison for each bifurcation, including the smaller Pcom and Acom bifurcations for angle features, is provided in Table \ref{table:bif_feature_comparison}.

\section{Discussions}
\subsection{Centerline Representations for Rich Morphometric Analysis}
The representation of the CoW as a centerline graph enables detailed morphometric analysis that is otherwise difficult to perform at scale. This work demonstrates several applications, including the automated classification of fetal-type PCA variants and the assessment of bifurcation symmetry and optimality relations. Notably, our analysis revealed subtle but consistent differences between imaging modalities, with MRA showing larger vessel radii compared to CTA. These insights would not be accessible through image- or segmentation-based inspection alone, highlighting the potential of centerline-based vascular analysis to support quantitative and reproducible assessments. Additional immediate applications include the detection of stenoses, hypoplastic vessels, and anatomical asymmetries in the CoW, all of which may influence the risk and outcomes of cerebrovascular pathologies. 

\subsection{Open Data for Benchmarking Vascular Centerline Extraction}
Evaluation of our U-Net–based centerline extraction pipeline demonstrates its ability to preserve vascular topology with high fidelity when combined with the A* skeleton connection algorithm. This topological accuracy is essential for downstream analyses and addresses a key limitation of conventional thinning-based skeletonization methods. Nonetheless, the current implementation has certain drawbacks, including relatively long inference times and limited evaluation on imperfect segmentation masks (see Section \ref{subsec:limitations} below). We believe that our publicly available centerline dataset and baseline algorithm provide a valuable foundation for benchmarking and further advancing learning-based extraction methods. An important aspect is that evaluation should not be limited to skeletonization performance alone, but should also consider the accuracy of derived anatomical quantities such as node location, variant classification, and morphometric features. To support this, we provide additional derived data alongside the centerline graphs to enable comprehensive and reproducible comparisons.

\subsection{Reliability of Morphometric Features Across Extraction Methods}
The comparative analysis of morphometric features derived from U-Net–predicted centerlines and manually verified reference graphs offers key insights into feature robustness. Features like segment radius and length showed strong agreement across methods, while others -- such as curvature and bifurcation angles -- exhibited greater variability and only moderate correlation, making them potentially less reliable for downstream clinical analysis. Importantly, we emphasize the need to assess not only the accuracy of absolute values but also the ability to preserve relative trends. This is relevant for features like tortuosity, where near-perfect correlation makes the feature valuable for clinical assessment despite variation in absolute measurements. Overall, our findings highlight the importance of validating the robustness and reliability of individual features in automated vascular analysis. 

\subsection{4.4 Limitations and Future Work}
\label{subsec:limitations}
While our method achieves strong topological accuracy and robust feature extraction, several limitations remain. First, the lack of clinically verified ground truth restricts validation of absolute morphometric feature values. 
Public datasets such as CROWN \citep{vos2025evaluation}, which include some validated feature measurements, could serve as useful reference for future validation efforts.
Second, the pipeline has so far only been implemented and tested on manually verified segmentation masks. A key next step will be to extend and evaluate it on imperfect mask predictions generated by segmentation models. 
Third, inference time for skeleton prediction remains relatively high due to heavy upsampling, model complexity and ensembling. This could be reduced by incorporating a slimmer architecture such as Skelite \citep{vargas2025skelite}.
Finally, some features — such as curvature and bifurcation angles — are inherently more sensitive to noise in the centerline and show greater variability across extraction methods. This variability should be carefully considered in downstream applications.

\section{Conclusion}
We present a publicly available dataset of anatomically labeled centerline graphs for the CoW, along with a baseline algorithm for centerline and feature extraction. By producing anatomically faithful centerline representations, our method enables reliable morphometric feature extraction for detailed quantitative vascular analysis beyond voxel-level segmentation. The dataset and pipeline provide a foundation for benchmarking and continued method development in cerebrovascular modeling and clinical research.

\section*{Data and Code Availability}
The training data, consisting of 250 labeled centerline graphs and derived quantities, is available in our public Zenodo repository: \href{https://zenodo.org/records/17358162}{https://zenodo.org/records/17358162}.

The code for the baseline algorithm is available on GitHub: \\
\href{https://github.com/fmusio/CoW_Centerline_Extraction/}{https://github.com/fmusio/CoW\textunderscore Centerline\textunderscore Extraction/}.

\section*{Acknowledgements}
This work was supported by the EU project GEMINI (funded by the EU Horizon Europe R\&I programme, Grant No. 101136438).
We thank Prof. Dr. med. Susanne Wegener for providing the medical images used in the TopCoW dataset, which form the basis of this work. We also thank PD Dr. med. Philippe Bijlenga for providing the fetal PCA ground truth annotations for TopCoW,  which were reused in this study. Furthermore, we thank Dr. med. Hakim Baazaoui and Julian Deseö for helpful discussions and clinical insights.


\FloatBarrier
\bibliographystyle{elsarticle-num} 
\bibliography{cas-refs}

\clearpage
\begin{center}
{\LARGE \textbf{Supplementary Material}}

\vspace{0.2cm}

{For the paper “\textit{Circle of Willis Centerline Graphs: A Dataset and Baseline Algorithm}”}

\vspace{0.5cm}
\end{center}
\FloatBarrier
\appendix
\section{Voreen Graph Processing Details}
\label{appendix:graph_extraction_details}
\subsection{Voreen Parameters for Centerline Extraction}
The centerline graphs are extracted using the Voreen tool \citep{drees2021scalable}: 
The extraction protocol consists of four stages -- skeletonization via topological thinning, topology extraction, voxel-branch assignement and feature extraction -- that are iteratively refined. A key parameter influencing the resulting graph structure is the scale-independent bulge size, which defines the minimum extent a protrusion must have from a parent vessel to be classified as a separate branch. For the task of CoW vessel graph extraction a bulge size of 1 was used. 
On overview of all parameter settings is presented in Table \ref{tab:voreen_parameters}.
\begin{table*}[ht!]
\caption{Overview of Voreen parameters for centerline extraction: A bulge size of 1 defines the minimum protrusion required for separate branch classification. The multi-class mask is binarized using a threshold of 0.5. Connectivity is determined using a 26-neighbourhood (n26), and surface smoothing is disabled. Cut-off regions are filtered based on a relative minimum bounding box diagonal of 0.05, with the absolute threshold disabled by setting it to 0mm.}
\centering
    \begin{tabular}{l|l}
        \hline
        \textbf{Voreen Parameter} & \textbf{Value} \\ \hline
        bulge size & 1 \\ \hline
        binarization threshold & 0.5 \\ \hline
        neighbourhood mode & n26 \\ \hline
        surface smoothing & False \\ \hline
        total min. bounding box diagonal & 0mm \\ \hline
        relative min. bounding box diagonal & 0.05 \\ \hline
    \end{tabular}
    \label{tab:voreen_parameters}
\end{table*}

\subsection{Post-processing of Extracted Graphs}
\label{appendix:graph_post_processing}
To correct topological artifacts introduced during skeletonization, we apply a structured post-processing pipeline consisting of the following steps:
\begin{itemize}
    \item Node extraction: Anatomically relevant graph nodes were identified based on known CoW topology. These nodes serve as anchors for correcting graph connectivity and guiding segment merging. A complete list of extracted nodes is provided in Table~\ref{table:nodes_per_segment}.
    \item Segment merging: Certain vessel segments, particularly the ACAs and Pcoms, were prone to topological errors when extracted from the full binarized mask. To mitigate this, these centerline segments were re-extracted from their respective single-class masks and merged with the main graph at the identified bifurcation nodes. 
    \item Rule-based edge removal: Remaining spurious edges, incorrect connections, and loops were removed using a set of anatomical rules. This step relied heavily on the previously extracted nodes to define valid segment boundaries and ensure topological correctness.
    \item Trimming and Smoothing: Centerline segments were trimmed at the degree-1 nodes and smoothed with a moving average filter with a window size of 5 to reduce voxel-level irregularities.
\end{itemize}

\begin{table*}[!ht]
\caption{List of extracted nodes for each vessel segment. The presence of specific nodes may vary depending on the anatomical variant of the CoW, and not all nodes are present in every centerline graph. There are three classes of nodes: start and end nodes (typically of degree 1), segment boundary nodes (degree 2) and bifurcation nodes (degree 3).}
\label{table:nodes_per_segment}
\centering
    \begin{tabular}{l|l|l}
        \hline
        \multicolumn{3}{c}{\bfseries CoW nodes per segment}\\
        \hline
        \bfseries Segment & \bfseries Labels & \bfseries Nodes \\
        \hline
        BA & 1 & BA start, BA bifurcation, R-PCA boundary, L-PCA boundary \\
        \hline
        R/L-PCA & 2, 3 & BA boundary, Pcom bifurcation, Pcom boundary, PCA end \\
        \hline
        R/L-ICA & 4, 6 & \makecell[l]{ICA start, Pcom bifurcation, Pcom boundary, ICA bifurcation,\\ACA boundary, MCA boundary} \\
        \hline
        R/L-MCA & 5, 7 & ICA boundary, MCA end \\
        \hline
        R/L-Pcom & 8, 9 & ICA boundary, PCA boundary \\
        \hline
        Acom & 10 & \makecell[l]{R-ACA boundary, L-ACA boundary, 3rd-A2 bifurcation,\\3rd-A2 boundary} \\
        \hline
        R/L-ACA & 11, 12 & ICA boundary, Acom bifurcation, Acom boundary, ACA end \\
        \hline
        3rd-A2 & 15 & Acom boundary, 3rd-A2 end \\
        \hline
    \end{tabular}
\end{table*}

\section{Feature Extraction Details}
\label{appendix:feature_extraction_details}
\subsection{Comparison of Radius Estimation Methods}
\label{appendix:radius_comparison}
\begin{figure}[!ht]
    \centering
    \includegraphics[width=0.8\textwidth]{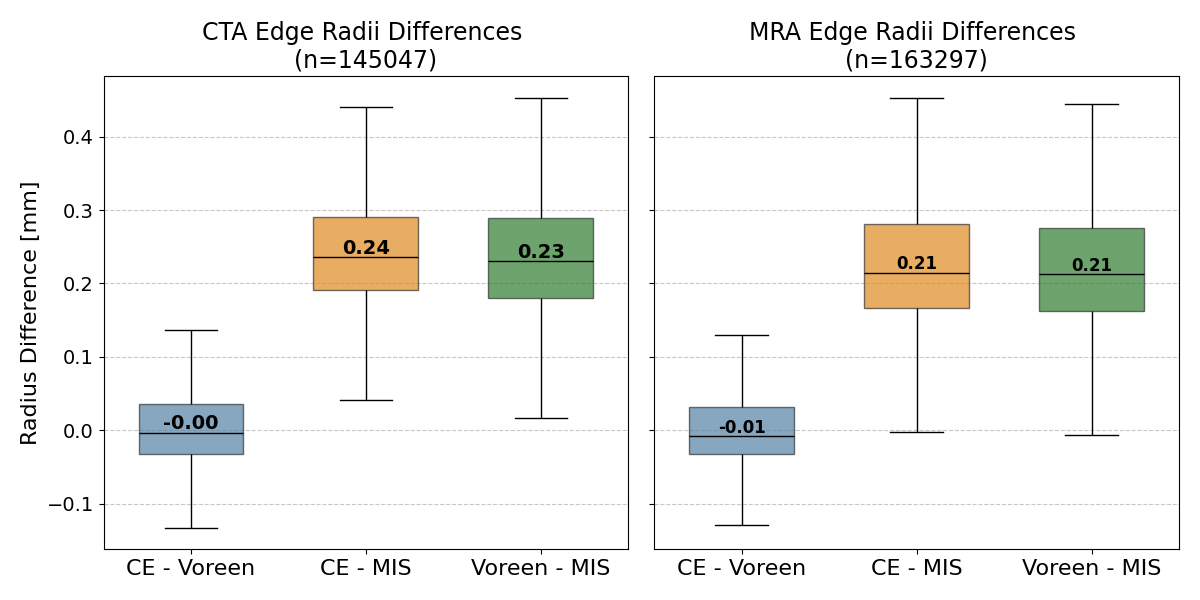}
    \caption{Pairwise differences in edge radii across the three estimation methods: Voreen average radius (Voreen), circle-equivalent (CE), and maximally inscribed sphere (MIS) radius for both CTA and MRA. For each CoW segment, radius differences were computed per edge. The boxplots summarize these differences across all segments and cases, with outliers excluded for clarity. Edges near bifurcations -- specifically those between bifurcation points and the boundaries of child vessels -- are excluded from the analysis, as they fall outside the defined anatomical segment boundaries. }
    \label{fig:edge_radii_diff}
\end{figure}

Besides the average edge radii provided by the Voreen tool and the circle-equivalent (CE) radii derived from cross-sectional area, we also compute the maximally inscribed sphere (MIS) radius as an additional edge attribute. Similar to the CE radius, this measure is based on a cross-sectional analysis along the centerline and is defined as the 10th percentile of distances from the centerline to the vessel surface.
Figure \ref{fig:edge_radii_diff} shows the differences in edge radii for the three estimation methods. The median of the pairwise differences between Voreen and CE radius estimates is very near zero, indicating excellent agreement between the methods. The MIS radii are consistently smaller, with median differences of 0.21–0.24mm, reflecting their more conservative definition.

\subsection{Morphometric Feature Extraction}
\label{appendix:morphometric_feature_extraction}
For segment features: In cases where communicating arteries are absent, we define the A1, P1, and C7 subsegments using their median lengths across the full TopCoW dataset: 15.57mm, 7.18mm, and 7.08mm, respectively. 
For segments directly adjacent to the CoW -- namely the BA, MCA, and 3rd-A2 segments, as well as the A2, P2, and C6 subsegments -- we limit the analysis to a maximum of 10 mm from the segment's origin. An exception is made for the C6 subsegment in CT scans, where the analysis is truncated at 5 mm due to the ICA entering the skull bone, rendering it invisible beyond that point.

\begin{figure}[!ht]
    \centering
    \includegraphics[width=1\textwidth]{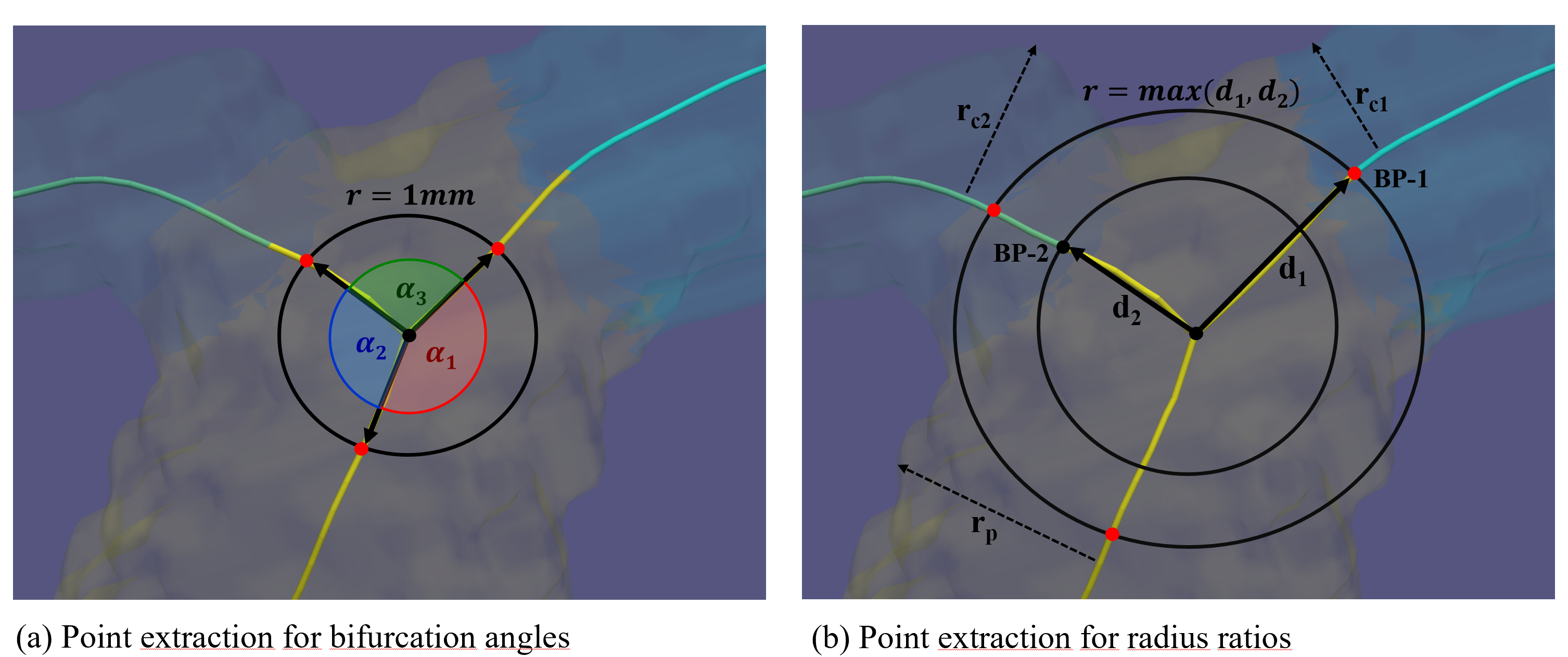}
    \caption{Equidistant point extraction for bifurcation feature computation. (a) Points (red) extracted at a distance of $r=1$mm from the bifurcation point are used to estimate the bifurcation angles between the parent and each child vessel ($\alpha_1$, $\alpha_2$) and between the two child vessels ($\alpha_3$). (b) Points (red) extracted at the maximum distance max($d_1, d_2$) from the bifurcation point to the segment boundary points are used to estimate the radius ratios and the bifurcation exponent. Vessel radii are sampled at these locations for the parent ($r_p$) and child vessels ($r_{c1}, r_{c2}$). The boundary points BP-1 and BP-2, marking the start of the child segments, are defined by the TopCoW annotations.}
    \label{fig:bif_features}
\end{figure}

For bifurcation features: To ensure robust angle measurements, we avoid using tangent vectors directly at the bifurcation point, as these can be sensitive to local noise or small variations. Instead, we extract points located 1mm away from the bifurcation, as illustrated in Figure \ref{fig:bif_features}(a), to obtain more stable directions.
For computing radius ratios and the bifurcation exponent in major bifurcations, we define the sampling location based on the labeled segment boundaries in the TopCoW annotations. Specifically, we determine the maximum distance from the bifurcation point to the start of each child vessel and sample the radii at this distance, as shown in Figure \ref{fig:bif_features}(b). To further stabilize these measurements, we average the radii over $n=3$ consecutive edges.  

\section{CoW Variant Classification}
\label{appendix:cow_variant_classification}
\begin{figure}[!ht]
    \centering
    \includegraphics[width=0.7\textwidth]{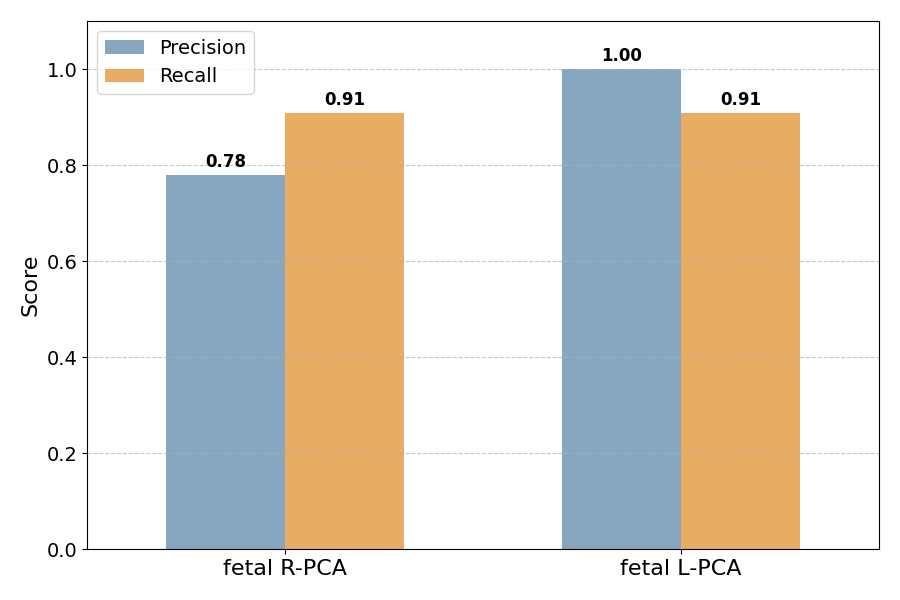}
    \caption{Fetal PCA classification performance in terms of precision and recall scores on a subset of 40 CTA images rated by a senior clinician. The comparison is made at the 25th percentile, where the Pcom diameter is at least 1.05 times larger than the P1 diameter.}
    \label{fig:fetal_eval}
\end{figure}

Building on the anterior (AV) and posterior (PV) variant classification introduced by TopCoW \citep{yang2023benchmarking}, we use ten binary variables (1 being present, 0 being absent) to describe the CoW variant:
\begin{itemize}
    \item Anterior: L-A1, Acom, 3rd-A2, R-A1
    \item Posterior: L-Pcom, L-P1, R-P1, R-Pcom
    \item Fetal: L-PCA, R-PCA
\end{itemize}
These variables can be automatically derived from the segmentation mask and the corresponding centerline graph with radius attributes, as demonstrated in the TopCoW summary paper.\\
For the communicating arteries and the 3rd-A2 segment, presence is determined by whether the corresponding label appears in the segmentation mask. In practice, a minimum size threshold (e.g., $\geq$30 voxels) can be applied to reduce false positives, especially when working with imperfect model outputs. 
For the A1 and P1 segments, presence is confirmed if the ACA and PCA labels, respectively, are present and connected to the ICA and BA, respectively. \\
To identify fetal PCA variants, we use the extracted centerline graphs and associated radii. Specifically, the radii along the Pcom and P1 segments are compared at the 25th percentile. If the Pcom diameter is at least slightly (1.05x) larger than the P1 diameter, the configuration is classified as a fetal PCA variant. To support the evaluation of the automated fetal PCA classification, a subset of 40 CTA images was independently annotated by a senior clinician through visual inspection as part of the TopCoW validation study. Using these expert labels as ground truth, the automated method achieved very high precision and recall values on the TopCoW annotations, as shown in Figure \ref{fig:fetal_eval}.

\section{Characteristics of the TopCoW Cohort}
\label{appendix:topcow_characteristics}

\subsection{Variant Distribution}
\label{appendix:variant_distribution}
The CoW variant distribution, as defined by their anterior variant (AV) and posterior variant (PV) graphs, is summarized in the TopCoW challenge supplementary material \citep{yang2023benchmarking}.
This section provides detailed statistics on missing arterial segments, fetal-type PCA variants, the presence of a 3rd-A2 segment (ACA trifurcation), and arterial fenestrations, with comparisons to values reported in the literature.

Table \ref{table:missing_segments_overall} summarizes the frequency of missing vessel segments and compares them to a study on $n=1864$ TOF-MRA from the Tromsø study by Hindenes et al. \citep{hindenes2020variations}. The prevalence of missing segments in the TopCoW dataset appears to align well with the findings reported by Hindenes et al. For the TopCoW MRA data, the chi-squared test reveals a statistically significant difference ($p<0.05$) only for the Acom and P1 segments. This discrepancy may be attributed to differences in annotation protocols: Hindenes et al. did not differentiate between missing and hypoplastic ($<1$mm) segments; i.e., when they refer to `missing' segments they mean `missing or hypoplastic'. Notably, when the TopCoW CTA data are also taken into account, the previously observed statistical significance is no longer present.
Furthermore, an asymmetry is observed between left and right missing Pcoms, with the left side being more frequently absent across all three datasets. This difference is statistically significant ($p<0.05$) in the Tromsø dataset. Finally, as noted in the main manuscript, missing segments are more frequently observed in CTA than in MRA. 
\begin{table*}[!ht]
\caption{The count and percentage of missing CoW segments are shown for the TopCoW CTA, TopCoW MRA, and the TOF-MRA from the Tromsø study as reported by Hindenes et al. For the Pcoms we distinguish between bilaterally and unilaterally missing segments. For A1 and P1 segments, left and right sides were combined and analyzed as a single group.}
\label{table:missing_segments_overall}
\centering
    \begin{tabular}{l|l|l|l}
        \hline
        \multicolumn{4}{c}{\bfseries CoW missing segments}\\
        \hline
        \bfseries Segment & \makecell[l]{\bfseries TopCoW CT\\\bfseries  (n=200)} & \makecell[l]{\bfseries  TopCoW MR\\\bfseries  (n=200)} & \makecell[l]{\bfseries  Tromsø MR\\\bfseries  (n=1864)} \\
        \hline
        Pcom bilat. & 82 (41.0\%) & 80 (40.0\%) & 720 (38.6\%)\\
        L-Pcom unilat. & 39 (19.5\%) & 37 (18.5\%) & 402 (21.6\%)\\
        R-Pcom unilat. & 28 (14.0\%) & 26 (13.0\%) & 257 (13.7\%)\\
        Acom & 34 (17.0\%) & 32 (16.0\%) & 420 (22.5\%)\\
        P1 & 26 (13.0\%) & 20 (10.0\%) & 330 (17.7\%)\\
        A1 & 11 (5.5\%) & 11 (5.5\%) & 61 (3.3\%)\\
        \hline
    \end{tabular}
\end{table*}

Table \ref{table:fetal_3rd-a2_frequency} presents the frequency of the fetal variant across the entire TopCoW dataset as well as the number of patients exhibiting a 3rd-A2 segment. Consistent with the trends observed in missing segment analysis above, the 3rd-A2 segment is more frequently identified in MRA, while the fetal variant is more commonly detected in CTA. Additionally, we compare the numbers to those from a study on $n=1739$ CTA scans conducted at Patras University Hospital by Zampakis et al. \citep{zampakis2015common}. The frequency of the unilateral fetal PCA variants in our dataset aligns well with the values reported by Zampakis et al. Similar to the unilaterally missing Pcoms above, an asymmetry is observed between the left and right fetal variants, with the left fetal PCA being more prevalent.  This difference is statistically significant ($p<0.05$) for the Patras dataset. 
Most notably, a chi-squared test indicates a highly significant difference ($p<0.00001$) in the frequency of the 3rd-A2 segment between the TopCoW cohort and the Patras dataset; however, the observed values remain within the broader prevalence range reported in the literature \citep{dimmick2009normal}. 
\begin{table*}[!ht]
\caption{The count and percentage of the fetal variant and the presence of a 3rd-A2 segment for the TopCoW CTA, TopCoW MRA, and the Patras hospital CTA as reported by Zampakis et al. For fetal variants, we differentiate between those that are present bilaterally and those that are present unilaterally.}
\label{table:fetal_3rd-a2_frequency}
\centering
    \begin{tabular}{l|l|l|l}
        \hline
        \multicolumn{4}{c}{\bfseries CoW fetal and 3rd-A2 variants}\\
        \hline
        \bfseries Variant & \makecell[l]{\bfseries TopCoW CT\\\bfseries (n=200)} & \makecell[l]{\bfseries TopCoW MR\\\bfseries (n=200)} & \makecell[l]{\bfseries Patras CT\\\bfseries (n=1739)} \\
        \hline
        fetal PCA bilat. & 16 (8.0\%) & 10 (5.0\%) & 77 (4.4\%) \\
        fetal L-PCA unilat. & 22 (11.0\%) & 21 (10.5\%) & 209 (12.0\%) \\
        fetal R-PCA unilat. & 16 (8.0\%) & 18 (9.0\%) & 113 (6.5\%) \\
        3rd-A2 & 21 (10.5\%) & 23 (11.5\%) & 26 (1.5\%)\\
        \hline
    \end{tabular}
\end{table*}

Finally, the prevalence of fenestrations in the Acom, A1, and P1 segments is very low, ranging between 0.5\% and 1.5\%, with consistently higher detection rates observed in the MRA modality. Reported values in the literature vary widely and appear to be strongly influenced by the imaging modality and its sensitivity to small vessel structures \citep{makowicz2013variants}.

\subsection{CoW Segment and Bifurcation Features}
\label{appendix:cow_features}
We report the median radius, length, and tortuosity for key CoW segments in Table \ref{table:cow_segment_features}, focusing exclusively on vessels strictly confined within the CoW. Segments are analyzed separately depending on whether the corresponding communicating artery (Acom or Pcom) is present. In cases where these arteries are absent, the A1, P1, and C7 subsegments are defined using median lengths as described in \ref{appendix:feature_extraction_details}, and their features are reported separately in gray. 
MRA exhibits a larger median radius for almost all segments, as already analyzed in the main manuscript. 
A statistically significant trend is observed for the P1 segment: in the absence of the Pcom, the P1 radius is larger and tortuosity is lower (Mann-Whitney U test, $p<0.05$). For the A1 segment, a slight trend toward a smaller radius is seen when the Acom is absent, reaching statistical significance only in CTA. 
\begin{table*}[!ht]
\caption{Median (Q1-Q3) values of segment features across various CoW segments reported for both CTA and MRA. For bilateral segments, left and right sides were combined and analyzed as a single group. The A1 segment was included once when the Acom was present (in black) and once when the Acom was absent (\textcolor{gray}{in gray}). Similarly, the P1 and C7 segments were included once when the Pcom was present (in black) and once when it was absent (\textcolor{gray}{in gray}). For absent communicating arteries we use the median lengths as definition of the A1, P1 and C7 segments.}
\label{table:cow_segment_features}
\centering
    \begin{tabular}{l|l|l|l|l}
        \hline
        \multicolumn{5}{c}{\bfseries CTA segment features}\\
        \hline
        \bfseries Segment & \bfseries Support & \makecell[l]{\bfseries Median radius} & \makecell[l]{\bfseries Length} & \makecell[l]{\bfseries Tortuosity} \\
        \hline
        PCA P1 & 143 & $1.06$ $(0.87-1.25)$ & $7.26$ $(5.86-9.73)$ & $0.11$ $(0.06-0.20)$\\
        \textcolor{gray}{PCA P1} & \textcolor{gray}{231} & \textcolor{gray}{$1.24$ $(1.11-1.32)$} & \textcolor{gray}{$7.16$ $(7.09-7.25)$} & \textcolor{gray}{$0.07$ $(0.05-0.11)$}\\
        ICA C7 & 169 & $2.00$ $(1.80-2.18)$ & $7.04$ $(6.46-7.89)$ & $0.06$ $(0.04-0.08)$\\
        \textcolor{gray}{ICA C7} & \textcolor{gray}{230} & \textcolor{gray}{$1.96$ $(1.79-2.13)$} & \textcolor{gray}{$7.07$ $(7.00-7.14)$} & \textcolor{gray}{$0.06$ $(0.04-0.08)$}\\
        Pcom & 169 & $0.86$ $(0.67-1.12)$ & $13.48$ $(11.75-15.07)$ & $0.17$ $(0.10-0.31)$\\
        Acom & 167 & $1.27$ $(1.06-1.49)$ & $2.65$ $(2.06-3.50)$ & $0.02$ $(0.01-0.05)$\\
        ACA A1 & 324 & $1.16$ $(1.02-1.30)$ & $15.63$ $(14.11-17.42)$ & $0.10$ $(0.07-0.16)$\\
        \textcolor{gray}{ACA A1} & \textcolor{gray}{68} & \textcolor{gray}{$1.24$ $(1.12-1.36)$} & \textcolor{gray}{$15.55$ $(15.50-15.64)$} & \textcolor{gray}{$0.10$ $(0.07-0.14)$}\\
        \hline
        \multicolumn{5}{c}{}\\
        \multicolumn{5}{c}{\bfseries MRA segment features}\\
        \hline
        \bfseries Segment & \bfseries Support  & \makecell[l]{\bfseries Median radius} & \makecell[l]{\bfseries Length} & \makecell[l]{\bfseries Tortuosity} \\
        \hline
        PCA P1 & 161 & $1.34$ $(1.04-1.51)$ & $7.09$ $(5.66-9.66)$ & $0.11$ $(0.06-0.22)$\\
        \textcolor{gray}{PCA P1} & \textcolor{gray}{223} & \textcolor{gray}{$1.51$ $(1.41-1.62)$} & \textcolor{gray}{$7.16$ $(7.11-7.25)$} & \textcolor{gray}{$0.07$ $(0.04-0.10)$}\\
        ICA C7 & 177 & $2.11$ $(1.96-2.25)$ & $7.19$ $(6.41-8.06)$ & $0.05$ $(0.04-0.07)$\\
        \textcolor{gray}{ICA C7} & \textcolor{gray}{223} & \textcolor{gray}{$2.10$ $(1.96-2.25)$} & \textcolor{gray}{$7.07$ $(6.99-7.15)$} & \textcolor{gray}{$0.06$ $(0.04-0.08)$}\\
        Pcom & 177 & $0.99$ $(0.74-1.16)$ & $12.46$ $(11.01-14.34)$ & $0.17$ $(0.09-0.27)$\\
        Acom & 171 & $1.24$ $(1.03-1.46)$ & $2.62$ $(1.99-3.54)$ & $0.03$ $(0.01-0.06)$\\
        ACA A1 & 328 & $1.36$ $(1.23-1.47)$ & $15.43$ $(13.88-17.06)$ & $0.09$ $(0.06-0.14)$\\
        \textcolor{gray}{ACA A1} & \textcolor{gray}{64} & \textcolor{gray}{$1.38$ $(1.26-1.47)$} & \textcolor{gray}{$15.54$ $(15.47-15.61)$} & \textcolor{gray}{$0.10$ $(0.06-0.13)$}\\
        \hline
    \end{tabular}
\end{table*}

Table \ref{table:cow_bifurcation_angle_features} presents all three bifurcation angles for both major and minor (shown in gray) CoW bifurcations.
At the BA bifurcation, the left and right branches exhibit symmetrical geometry, with parent-child angles being nearly equal.
For the ICA bifurcations, the MCA consistently forms a wider angle with the ICA -- approximately 25\% larger -- compared to the ACA.
Across modalities, bifurcation angles are generally consistent. Statistically significant differences (Wilcoxon signed-rank test, $p<0.05$) are observed only for the R-PCA/L-PCA and ICA/MCA angles, both of which are slightly larger in MRA.
\begin{table*}[!ht]
\caption{Median (Q1-Q3) values of bifurcation angle features across all major and minor (\textcolor{gray}{in gray}) CoW bifurcations, reported separately for CTA and MRA. For each bifurcation listed in row order, the parent vessel (p) and child vessels (c1, c2) are as follows: (p) BA, ICA, PCA P1, ICA C6, ACA A1; (c1) R-PCA, MCA, PCA P2, ICA C7, ACA A2; (c2) L-PCA, ACA, Pcom, Pcom, Acom.}
\label{table:cow_bifurcation_angle_features}
\centering
    \begin{tabular}{l|l|l|l|l}
        \hline
        \multicolumn{5}{c}{\bfseries CTA bifurcation angle features} \\
        \hline
        \bfseries Bifurcation & \bfseries Support & \makecell[l]{\bfseries Angle(p,c1)} & \makecell[l]{\bfseries Angle(p,c2)} & \makecell[l]{\bfseries Angle(c1,c2)} \\
        \hline
        BA & 173 & $122.4$ $(110.3-134.7)$ & $127.9$ $(113.2-139.7)$ & $106.3$ $(93.3-116.9)$ \\
        ICA & 386 & $133.0$ $(123.0-142.7)$ & $107.6$ $(97.0-116.7)$ & $114.7$ $(104.0-127.4)$ \\
        \textcolor{gray}{Pcom PCA} & \textcolor{gray}{143} & \textcolor{gray}{$150.7$ $(140.3-158.5)$} & \textcolor{gray}{$98.1$ $(89.5-107.6)$} & \textcolor{gray}{$109.5$ $(100.2-119.2)$} \\
        \textcolor{gray}{Pcom ICA} & \textcolor{gray}{167} & \textcolor{gray}{$158.8$ $(151.6-165.6)$} & \textcolor{gray}{$97.6$ $(88.5-107.7)$} & \textcolor{gray}{$102.7$ $(92.1-111.5)$} \\
        \textcolor{gray}{Acom} & \textcolor{gray}{322} & \textcolor{gray}{$154.4$ $(145.9-161.1)$} & \textcolor{gray}{$102.2$ $(91.5-111.5)$} & \textcolor{gray}{$102.4$ $(91.5-113.2)$} \\
        \hline
        \multicolumn{5}{c}{}\\
        \multicolumn{5}{c}{\bfseries MRA bifurcation angle features}\\
        \hline
        \bfseries Bifurcation & \bfseries Support & \makecell[l]{\bfseries Angle(p,c1)} & \makecell[l]{\bfseries Angle(p,c2)} & \makecell[l]{\bfseries Angle(c1,c2)} \\
        \hline
        BA & 181 & $120.7$ $(109.4-134.5)$ & $125.8$ $(110.0-137.3)$ & $109.1$ $(99.4-119.3)$ \\
        ICA & 387 & $134.8$ $(125.5-144.5)$ & $107.5$ $(97.9-117.7)$ & $113.4$ $(104.2-124.4)$ \\
        \textcolor{gray}{Pcom PCA} & \textcolor{gray}{158} & \textcolor{gray}{$152.5$ $(142.9-161.4)$} & \textcolor{gray}{$97.0$ $(89.5-106.4)$} & \textcolor{gray}{$105.8$ $(96.2-118.0)$} \\
        \textcolor{gray}{Pcom ICA} & \textcolor{gray}{176} & \textcolor{gray}{$158.7$ $(151.7-164.7)$} & \textcolor{gray}{$100.2$ $(91.5-109.8)$} & \textcolor{gray}{$102.0$ $(92.2-109.7)$} \\
        \textcolor{gray}{Acom} & \textcolor{gray}{330} & \textcolor{gray}{$153.9$ $(147.0-160.5)$} & \textcolor{gray}{$103.0$ $(92.5-113.9)$} & \textcolor{gray}{$101.3$ $(91.9-111.1)$} \\
        \hline
    \end{tabular}
\end{table*}

\section{Baseline Algorithm Details}
\label{appendix:baseline_details}
\subsection{Implementation Details}
\label{appendix:baseline_implementation_details}
Skeleton prediction is performed using a 3D U-Net implemented via the nnU-Net framework \citep{ronneberger2015u, isensee2023nnUNet}. Due to the severe class imbalance between skeleton and background voxels, we use a composite loss function that combines weighted Focal Loss and Tversky Loss, with a weight ratio of 2:1, respectively. The Focal Loss is hereby defined as
\begin{equation}
FL = -w_{f}(1-p)^\gamma log(p)-w_{b}p^\gamma log(1-p)
\end{equation}
where $w_{f}=0.75$, $w_{b}=0.25$ and $\gamma=2$. This loss emphasizes hard-to-classify foreground voxels and reduces the influence of easy background predictions.\\ 
The Tversky Loss is given by
\begin{equation}
    TI=\frac{TP}{TP+\alpha FP+\beta FN}
\end{equation}
where $\alpha=0.5$, $\beta=0.75$, placing greater weight on false negatives to improve recall of thin structures.\\
Training is conducted on 250 multi-class masks from the TopCoW training dataset, resampled to [0.25, 0.25, 0.25] mm spacing. Binary skeletons derived from the Voreen-extracted centerline graphs serve as targets. We use 5-fold cross-validation, training for 500 epochs per fold with deep supervision disabled and default data augmentation (excluding mirroring).

To reconnect fragmented skeletons, we first assign anatomical labels from the original segmentation mask to the binary predicted skeleton. We then apply a two-stage reconnection strategy: first within the same anatomical label, then across neighboring labels to ensure global CoW connectivity.
We use the A$^*$ algorithm to find optimal paths between disconnected endpoints. The cost function is defined as
\begin{equation}
    F(n_{i}, n_{f})=w_{1}d(n_{i}, n_{f}) - w_{2}f_d(n_{i})
\end{equation}
where $n_{i}$ and $n_f$ denote the start and target nodes, $d(n_{i}, n_{f})$ is the Euclidean distance between nodes, and $f_d(n_{i})$ is the distance from the vessel boundary, encouraging paths to remain centered. We use weights $w_1=1$ and $w_2=2$ in practice.

\subsection{Feature Comparison}
A detailed comparison of the different segment and bifurcation features obtained from the Voreen-extracted and the U-Net-predicted skeletons is provided in Table \ref{table:segment_feature_comparison} and Table \ref{table:bif_feature_comparison} respectively. For each feature, we report the median relative error (MedRE) and the Pearson correlation coefficient (Pearson r).\\
Except for curvature, the Acom consistently shows a much higher median MedRE across segment features, which led to its exclusion from the main analysis in Section \ref{subsubsec:cow_comparison}. For tortuosity, MedRE tends to increase with shorter segments, but remains reasonably low ($\sim$10\%) for the P1, Pcom, and A1 segments. Compared to these segments, length and tortuosity estimates for the C7 segment show significantly higher MedRE. Finally, Pearson r values are very high for all features except curvature, indicating strong linear trend alignment.\\
For bifurcation angles and exponent, the ICA bifurcations perform better than the BA bifurcation, with both lower MedRE and higher Pearson r for these features, possibly due to greater support. Interestingly, the minor communicating artery bifurcations exhibit even lower MedRE and substantially higher Pearson r for the bifurcation angles than the major BA and ICA bifurcations, suggesting better reliability in terms of both absolute values and linear trend alignment.
\label{appendix:baseline_feature_comparison}
\begin{table*}[!ht]
\caption{Comparison of segment features extracted from the Voreen-based and U-
Net–predicted centerline graphs across different CoW segments, evaluated by median relative error (MedRE) and Pearson correlation coefficient (Pearson r). For bilateral segments, left and right sides are combined and analyzed as a single group. No distinction is made between features extracted from CTA- and MRA-derived masks. Segments marked with a `*' were included only when the respective communicating artery was present. For cases with absent communicating arteries, median segment lengths are used to define the A1, P1, and C7 segments.}
\label{table:segment_feature_comparison}
\centering
\resizebox{1\textwidth}{!}{
    \begin{tabular}{l|l|l|l|@{\hskip 1pt}l|l|l|l|l}
        \cline{1-4} \cline{5-9}
        \multicolumn{4}{c|}{} & & \multicolumn{4}{c}{} \\
        \multicolumn{4}{c|}{\makecell[c]{\textbf{Median radius}}} & & \multicolumn{4}{c}{\makecell[c]{\textbf{Volume}}} \\
        \cline{1-4} \cline{5-9}
        \textbf{Segment} & \textbf{Support} & \makecell[l]{\bfseries MedRE (Q1-Q3)\\\bfseries [\%]} & \textbf{Pearson r} & & \textbf{Segment} & \textbf{Support} & \makecell[l]{\bfseries MedRE (Q1-Q3)\\\bfseries [\%]} & \textbf{Pearson r}  \\
        \cline{1-4} \cline{5-9}
        BA & 140 & $0.44$ $(0.18-0.81)$ & 1.00 & & BA & 140 & $1.94$ $(0.80-3.39)$ & 0.99 \\
        PCA P1 & 268 & $0.63$ $(0.27-1.23)$ & 1.00 & & PCA P1 & 268 & $1.99$ $(0.85-3.38)$ & 0.97 \\
        ICA C7 & 280 & $0.54$ $(0.26-1.00)$ & 1.00 & & ICA C7 & 280 & $2.12$ $(0.92-3.98)$ & 0.97 \\
        MCA & 280 & $0.46$ $(0.18-0.86)$ & 1.00 & & MCA & 280 & $1.42$ $(0.71-2.65)$ & 0.99 \\
        Pcom & 101 & $0.90$ $(0.32-1.60)$ & 1.00 & & Pcom & 101 & $2.26$ $(0.95-4.99)$ & 1.00 \\
        Acom & 115 & $1.82$ $(1.01-4.07)$ & 0.99 & & Acom & 115 & $9.15$ $(4.49-20.33)$ & 0.96 \\
        ACA A1 & 278 & $0.59$ $(0.22-1.04)$ & 1.00& & ACA A1 & 278 & $1.71$ $(0.81-3.12)$ & 1.00\\
        3rd-A2 & 13 & $0.77$ $(0.48-0.93)$ & 1.00& & 3rd-A2 & 13 & $0.78$ $(0.49-1.58)$ & 1.00\\
        \cline{1-4} \cline{5-9}
        \multicolumn{4}{c|}{} & & \multicolumn{4}{c}{} \\
        \multicolumn{4}{c|}{\makecell[c]{\textbf{Tortuosity}}} & & \multicolumn{4}{c}{\makecell[c]{\textbf{Curvature}}} \\
        \cline{1-4} \cline{5-9}
        \textbf{Segment} & \textbf{Support} & \makecell[l]{\bfseries MedRE (Q1-Q3)\\\bfseries [\%]} & \textbf{Pearson r} & & \textbf{Segment} & \textbf{Support} & \makecell[l]{\bfseries MedRE (Q1-Q3)\\\bfseries [\%]} & \textbf{Pearson r}  \\
        \cline{1-4} \cline{5-9}
        BA & 140 & $34.06$ $(15.57-52.16)$ & 0.94 & & BA & 140 & $37.15$ $(24.43-48.38)$ & 0.40 \\
        PCA P1 & 268 & $12.34$ $(6.85-21.96)$ & 0.99 & & PCA P1 & 268 & $22.54$ $(11.25-35.50)$ & 0.62 \\
        ICA C7 & 280 & $18.75$ $(8.90-28.69)$ & 0.86 & & ICA C7 & 280 & $28.81$ $(14.89-40.90)$ & 0.45 \\
        MCA & 280 & $22.57$ $(10.38-38.96)$ & 0.97 & & MCA & 280 & $39.05$ $(24.96-48.65)$ & 0.44 \\
        Pcom & 101 & $9.06$ $(3.95-14.04)$ & 0.98 & & Pcom & 101 & $17.55$ $(11.07-27.18)$ & 0.66 \\
        Acom & 115 & $46.67$ $(19.29-72.73)$ & 0.94 & & Acom & 115 & $25.85$ $(10.88-47.87)$ & 0.54 \\
        ACA A1 & 278 & $10.37$ $(5.68-15.68)$ & 0.99 & & ACA A1 & 278 & $29.40$ $(19.12-37.25)$ & 0.57 \\
        3rd-A2 & 13 & $24.32$ $(9.30-33.33)$ & 1.00 & & 3rd-A2 & 13 & $35.64$ $(9.71-47.75)$ & 0.35 \\
        \cline{1-4} \cline{5-9}
        \multicolumn{4}{c|}{} \\
        \multicolumn{4}{c|}{\makecell[c]{\textbf{Length}}} \\
        \cline{1-4}
        \textbf{Segment} & \textbf{Support} & \makecell[l]{\bfseries MedRE (Q1-Q3)\\\bfseries [\%]} & \textbf{Pearson r} \\
        \cline{1-4}
        PCA P1* & 89 & $2.65$ $(1.23-5.13)$ & 0.99 \\
        ICA C7* & 101 & $4.48$ $(1.97-6.95)$ & 0.94 \\
        Pcom & 101 & $1.83$ $(0.98-3.30)$ & 0.98 \\
        Acom & 115 & $8.70$ $(3.62-19.45)$ & 0.95 \\
        ACA A1* & 226 & $1.54$ $(0.82-2.66)$ & 0.99 \\
        \cline{1-4}
    \end{tabular}
}
\end{table*}

\begin{table*}[!ht]
\caption{Comparison of bifurcation features extracted from the Voreen-based and U-
Net–predicted centerline graphs across different CoW bifurcations, evaluated by median relative error (MedRE) and Pearson correlation coefficient (Pearson r). For bilateral segments, left and right sides are combined and analyzed as a single group. No distinction is made between features extracted from CTA- and MRA-derived masks. For each bifurcation, all three bifurcation angles and the three individual radius ratios between the parent and child vessels were included in the analysis. The radius sum and area sum ratios were calculated as the parent vessel's value divided by the sum of the corresponding child vessel values.}
\label{table:bif_feature_comparison}
\centering
\resizebox{1\textwidth}{!}{
    \begin{tabular}{l|l|l|l|@{\hskip 1pt}l|l|l|l|l}
        \cline{1-4} \cline{5-9}
        \multicolumn{4}{c|}{} & & \multicolumn{4}{c}{} \\
        \multicolumn{4}{c|}{\makecell[c]{\textbf{Individual ratios}}} & & \multicolumn{4}{c}{\makecell[c]{\textbf{Radius sum ratio}}} \\
        \cline{1-4} \cline{5-9}
        \textbf{Bifurcation} & \textbf{Support} & \makecell[l]{\bfseries MedRE (Q1-Q3)\\\bfseries [\%]} & \textbf{Pearson r} & & \textbf{Bifurcation} & \textbf{Support} & \makecell[l]{\bfseries MedRE (Q1-Q3)\\\bfseries [\%]} & \textbf{Pearson r}  \\
        \cline{1-4} \cline{5-9}
        BA & 378 & $2.51$ $(1.04-5.41)$ & 0.98 & & BA & 126 & $2.25$ $(0.99-4.68)$ & 0.95 \\
        ICA & 819 & $2.26$ $(0.98-4.27)$ & 0.98 & & ICA & 273 & $1.83$ $(0.88-3.37)$ & 0.94 \\
        \cline{1-4} \cline{5-9}
        \multicolumn{4}{c|}{} & & \multicolumn{4}{c}{} \\
        \multicolumn{4}{c|}{\makecell[c]{\textbf{Area sum ratio}}} & & \multicolumn{4}{c}{\makecell[c]{\textbf{Bifurcation exponent}}} \\
        \cline{1-4} \cline{5-9}
        \textbf{Bifurcation} & \textbf{Support} & \makecell[l]{\bfseries MedRE (Q1-Q3)\\\bfseries [\%]} & \textbf{Pearson r} & & \textbf{Bifurcation} & \textbf{Support} & \makecell[l]{\bfseries MedRE (Q1-Q3)\\\bfseries [\%]} & \textbf{Pearson r}  \\
        \cline{1-4} \cline{5-9}
        BA & 126 & $4.49$ $(1.90-8.89)$ & 0.95 & & BA & 126 & $7.14$ $(2.76-14.29)$ & 0.63 \\
        ICA & 273 & $3.71$ $(1.66-6.90)$ & 0.94 & & ICA & 273 & $6.19$ $(2.63-12.12)$ & 0.81 \\
        \cline{1-4} \cline{5-9}
        \multicolumn{4}{c|}{} \\
        \multicolumn{4}{c|}{\makecell[c]{\textbf{Bifurcation angles}}} \\
        \cline{1-4}
        \textbf{Bifurcation} & \textbf{Support} & \makecell[l]{\bfseries MedRE (Q1-Q3)\\\bfseries [\%]} & \textbf{Pearson r} \\
        \cline{1-4}
        BA & 378 & $8.94$ $(4.38-17.03)$ & 0.55 \\
        ICA & 819 & $8.15$ $(3.61-14.18)$ & 0.64 \\
        Pcom PCA & 261 & $6.14$ $(1.98-12.12)$ & 0.91 \\
        Pcom ICA & 297 & $6.35$ $(2.60-14.41)$ & 0.88 \\
        Acom & 672 & $7.22$ $(3.30-13.51)$ & 0.87 \\
        \cline{1-4}
    \end{tabular}
}
\end{table*}





\end{document}